\def\year{2022}\relax
\documentclass[letterpaper]{article} 
\usepackage{aaai22}  
\usepackage{times}  
\usepackage{helvet}  
\usepackage{courier}  
\usepackage[hyphens]{url}  
\usepackage{graphicx} 
\usepackage{natbib}  
\usepackage{caption} 
\DeclareCaptionStyle{ruled}{labelfont=normalfont,labelsep=colon,strut=off} 
\usepackage{algorithm}
\usepackage{algorithmic}
\usepackage[latin1]{inputenc}
\usepackage{amsmath}
\usepackage{amsfonts,amssymb}
\usepackage{newfloat}
\usepackage{listings}
\usepackage{cite}
\floatstyle{ruled}
\newfloat{listing}{tb}{lst}{}
\floatname{listing}{Listing}
\title{How Well Do Self-Supervised Methods Perform in Cross-Domain \\
       Few-Shot Learning?}
 \affiliations{
    \textsuperscript{\rm 1}Artificial Intelligence Research Institute, Zhejiang Lab, China\\
     \textsuperscript{\rm 2}College of Computer and Information Engineering, Zhejiang Gongshang University, China\\
     yiyi.zhang93@outlook.com, zhengyinghit@outlook.com

 }
 \author{
     Yiyi Zhang\textsuperscript{\rm 1}\equalcontrib,
     Ying Zheng\textsuperscript{\rm 1}\equalcontrib,
     Xiaogang Xu\textsuperscript{\rm 2}\thanks{Corresponding author.},
     Jun Wang\textsuperscript{\rm 1}
     \\
 }

\begin{document}

\maketitle

\begin{abstract}
Cross-domain few-shot learning (CDFSL) remains a largely unsolved problem in the area of computer vision, while self-supervised learning presents a promising solution. Both learning methods attempt to alleviate the dependency of deep networks on the requirement of large-scale labeled data. Although self-supervised methods have recently advanced dramatically, their utility on CDFSL is relatively unexplored. In this paper, we investigate the role of self-supervised representation learning in the context of CDFSL via a thorough evaluation of existing methods. It comes as a surprise that even with shallow architectures or small training datasets, self-supervised methods can perform favorably compared to the existing SOTA methods. Nevertheless, no single self-supervised approach dominates all datasets indicates that existing self-supervised methods are not universally applicable. In addition, we find that representations extracted from self-supervised methods exhibit stronger robustness than the supervised method. Intriguingly, whether self-supervised representations perform well on the source domain has little correlation with their applicability on the target domain. As part of our study, we conduct an objective measurement of the performance for six kinds of representative classifiers. The results suggest Prototypical Classifier as the standard evaluation recipe for CDFSL.
\end{abstract}

\section{Introduction}
We humans can refer to our past practical experience and quickly condense new concepts from limited data. Few-shot learning \cite{c:human} seeks to imitate this process by learning representations from base classes that can be well generalized to novel classes where only a few samples are available. Typically, few-shot learning consists of two stages: meta-learning and meta-testing. During meta-learning, a system is trained on an abundance of base classes in order to learn well with few labeled examples from that particular domain. In the meta-test phase, there is a set of novel classes, consisting of a handful of labeled and testing examples per class, used to evaluate the trained system. However, learning feature encoders from base classes might discard semantic information that is irrelevant for base classes but essential for novel classes, thus may lead to over-fitting.

One way to mitigate this over-fitting problem is to leverage embedding learning methods that do not use class labels, such as self-supervised learning \cite{c:sslk}. It was originally proposed to alleviate the challenging demand for large amounts of annotated data by learning about statistical regularities within images. Recently, \cite{c:rfs} has raised hopes that self-supervised methods can approach and even surpass the ubiquitous annotation-intensive paradigm of supervised learning in the scenario of few-shot learning. Despite recent progress, their applicability to cross-domain few-shot learning is unclear. From this perspective, a core issue is how well self-supervised methods perform in cross-domain few-shot learning. A thorough investigation into this issue will facilitate the expansion and practical implementation of CDFSL within the domains of intelligent vision \cite{zheng2020deep,zheng2021sketch,zhang2025leaf} and robotics \cite{zheng2024survey}.

Driven by the great value and challenge of CDFSL \cite{c:broader}, we put forward the following thoughts to further dissect our core question. 1) \emph{To what extent self-supervised features can transfer directly to uncharted domains?} With the continuous emergence of works on few-shot learning by meta-learning, multi-task learning, supervised and self-supervised methods, there is still a huge lack of thorough comparison of these methods on cross-domain few-shot learning. 2) \emph{Which one is the most practically effective for cross-domain few-shot learning among the plethora of self-supervised methods in the market?} It can \emph{not} be easily appraised given the limited commonality of training and evaluation conditions reported by each method. One thing important to know especially, most of the state-of-the-art self-supervised approaches are trained with deep networks (e.g. ResNet50) while network architectures for few-shot learning are usually shallow (e.g. ResNet10). Whether self-supervised methods with shallow networks are superior to other methods remains to be tested. 3) \emph{Given various choices of classifiers, which one is the best and how to measure their performance objectively?} On the one hand, a framework is to be established in an effort to methodically select a reliable classifier w.r.t its capability, stability, and velocity. On the other hand, we aim to be aware of which classifier can fairly and efficiently evaluate each embedding, such that it can be applied as a standard evaluation recipe for cross-domain few-shot learning.

To answer the above questions, we conduct a big body of practical experiments on the performance of up-to-date pre-trained embeddings. These experiments include aspects from diverse training protocols, different backbones, and various classifiers. This is a comprehensive evaluation that has been missing in the literature until now. In particular, while we hope that embeddings assessed with protruding performance by one specific classifier will also perform well on alternative ones (\emph{a good embedding is all you need?}), this conjecture has never been systematically tested for cross-domain few-shot learning. Extensive experimental results uncover several impressive insights, in which the most crucial are summarized as follows:
\begin{itemize}
\item The most advanced self-supervised methods surpass existing state-of-the-art methods on benchmarks, even by a considerable margin under certain cases. However, no single self-supervised method dominates all datasets.
\item Self-supervised methods show greater advantages and deliver more robust representations than the supervised method, especially when the training data is smaller or the task is more challenging.
\item The performance of self-supervised methods on the source domain has little correlation with their applicability on the target domain. This suggests that an appropriate way to select the best models during training for the cross-domain settings has not yet been realized.
\item We define formal evaluation criteria for classifiers on cross-domain few-shot learning. As a result, Prototypical Classifier is recognized as the most reliable one.
\end{itemize}

\section{Related work}
\subsection{Few-shot Learning (FSL)}
Few-shot learning has attracted considerable attention with a wide range of promising methods, including optimization-based meta-learners \cite{c:maml,c:optim,c:convex}, which learn to adapt quickly on tasks with only a few training samples available; metric learning based protocols that learn a distance metric between a query and a bunch of support images \cite{c:proto,c:relation,c:match}; approaches try to synthesize more training data or features to alleviate the data insufficiency problem \cite{c:freelunch,c:wang}.

Though the sophisticated meta-learning based methods achieve SOTA performance in FSL, \cite{c:closer,c:broader} claim that existing algorithms may perform undesirably and even underperform to the traditional supervised learning method, when there is a obvious domain gap between training and testing tasks. To this end, we aim to seek help from self-supervised learning to moderate the over-fitting problem.
\subsection{Self-supervised Learning (SSL)}
Since the image itself contains structural information that can be utilized to design pretext tasks, SSL takes advantage of the surrogate supervision signal provided by the pretext task to perform feature learning. The main challenge here is how to construct a useful pretext task. A common paradigm is to pre-train on ImageNet by predicting the color \cite{c:color}, rotation \cite{c:rotation}, relative patch location \cite{c:patch}, instance discrimination \cite{c:moco,c:insdis} and clusters \cite{c:cluster1,c:cluster2}.

In recent works, a study about embedding networks on few-shot learning \cite{c:rfs} indicates that a self-supervised resnet50 is competitive to the supervised one, showing the potential of SSL in the scenario of few-shot learning. \cite{c:bf3s} introduce a multi-task paradigm by weaving self-supervised learning into the training objective of few-shot learning to boost the ability of the latter to adapt to new tasks quickly. A work \cite{c:su} show that attaching self-supervised tasks using data across domains can boost the performance for conventional few-shot learning. We further investigate these SSL methods in the cross-domain few-shot learning setting.
\subsection{Cross-domain Few-shot Learning (CDFSL)}
For cross-domain few-shot learning, the base and novel classes are drawn from different domains with disjoint class labels. In prior works, \cite{c:tseng} propose the feature-wise transformation layers to strengthen the robustness of the metric functions. \cite{c:sun} elaborate an explanation-guided training strategy to prevent the feature extractor from over-fitting to specific classes. While they are proposed only for the realm of natural images, but those who improve CDFSL on MiniImage to CUB do not perform as well as in unnatural target domains (e.g., training on the MiniImageNet classes and testing on the ChestX classes).

To further elevate the CDFSL performance under sharp domain shifts, \cite{c:taskaug} focus on improving the robustness of various inductive bias through task augmentation. \cite{c:cai} apply graph neural network (GNN) as graph-based convolutions to learn a domain-agnostic embedding by fitting to pre-softmax classification scores from finetuned feature encoders. However, common to these meta-learning based works is that they need to be pre-trained on the base classes in a supervised way beforehand, so we exclude them in our experiments for a fair comparison. Recently, SSL is also introduced into CDFSL \cite{c:amdim,c:prototrans}, revealing the strong generalization ability of SSL trained feature extractors. Note that we primarily focus on inductive learning, in which the information from testing data is not utilized since testing data is generally unavailable in the open world.

\section{Preliminaries}
\subsection{Problem Formulation}
Formally, a few-shot task can be defined as $\tau=(\tau_s, \tau_q)$, where $\tau_s$ and $\tau_q$ represent a support set and a query set respectively. Typically, the query set $\tau_q$ consists of samples from the same classes as $\tau_s$. A few-shot classification task is called $C$-way $K$-shot if the support set $\tau_s$ contains $C$ classes with $K$ samples. With the support set $\tau_s\in\left\{\tau_i\right\}^{C\times K}_{i=1}$, our goal is to classify the samples from the query set $\tau_q\in\left\{\tau_i\right\}^{Q}_{j=1}$ into one of the $C$ classes. We produce source tasks from base classes and target tasks from novel classes respectively. In general, the target tasks $\widetilde{\tau}$ are assumed to come from the source task distribution $\Phi_s$. However, in this work, we consider FSL under domain shifts. Which is to say, base classes and novel classes come from different domains. Concretely, we concentrate on the single domain methods, where base classes from only one domain are available.
\subsection{Evaluated Methods}
We compare embeddings trained with different protocols, including the supervised method, self-supervised methods, and the multi-task method. The classifiers we use to evaluate embedding networks include: Logistic Regression (LR) \cite{c:lr}, SVM \cite{c:svm}, Cosine Classifier (CC) \cite{c:dynamic}, Nearest Neighbor (NN) \cite{c:knn}, Prototypical Classifier (Proto) \cite{c:bf3s} and Linear Classifier (LC). For LC, the attached fully connected layer is trained from scratch each time, when a new support set is in process. Specifically, the obtained features from each embedding network will be normalized before fed to the classifiers, except for LC.

\noindent\textbf{Supervised method}\quad For the supervised way, the network is trained from base classes and validated by the validation set from source distribution $\Phi_s$. In order to measure the performance of each method equally, all networks are pre-trained from scratch. We fix the pre-trained backbone for all methods, then attach various classifiers to estimate respectively how well each feature extractor transfers to tasks from the target task distribution $\Phi_t$.

\noindent\textbf{Meta-learning methods}\quad In few-shot learning, the meta-training stage(known as episode training) uses source data to mimic the meta-test scenario, such that the model trained on task $\tau$ can quickly adapt to task $\widetilde{\tau}$. We inherit the meta-learning methods used from \cite{c:broader}: MAML \cite{c:maml}, ProtoNet \cite{c:proto}, relationNet \cite{c:relation} and MetaOpt \cite{c:convex}. These methods implicitly assume that task $\tau$ share the same distribution with task $\widetilde{\tau}$, so the task-agnostic knowledge can be leveraged for fast learning on novel classes. However, it poses a great challenge for them to transfer well to novel classes , where the distribution is of a great gap to base classes.

\noindent\textbf{Multi-task learning method}\quad We train the feature encoder $\mathcal{F}_\theta$ with both annotated data and non-annotated data in a multi-task setting. For the SSL branch, we consider two pretext tasks motivated by the recent work \cite{c:scaling}:

- \textit{predicting the rotation incurred by the image}, we follow the method \cite{c:rotation} where each image $x$ is rotated by an angle $\Lambda\in \left\{0^{\circ},90^{\circ},180^{\circ},270^{\circ}\right\}$ to obtain $\hat{x}$ and the target label $\hat{y}$ is the index of the angle. Based on the encoded feature $\mathcal{F}_\theta(\hat{x})$, the rotation classifier $\mathcal{R}_\phi$ attempts to predict the rotation class. The self-supervised loss of this task is defined as:
\begin{equation}
\begin{split}
        L_{self}(\theta,\phi;\hat{\mathcal{X}}) = -log \sum \limits_{\forall \hat{x} \in \hat{\mathcal{X}}} \hat{y}(\mathcal{R}_\phi(\mathcal{F}_\theta(\hat{x})))
    \end{split}
    \label{eq:eqname1}
\end{equation}
where $\hat{\mathcal{X}}$ is the rotated training image set.

- \textit{predicting the relative patch location}, here the input image is tiled into 3x3 patches and paired randomly to obtain the input pair $\bar{x}$ according to the process outlined in \cite{c:patch}. The target label $\bar{y}$ is the index of the relative location of the first patch w.r.t the first one in one input pair. We use a fully connected network $\mathcal{P}_\phi(\cdot)$ to predict the target label. Accordingly, the self-supervised loss is formulated as:
\begin{equation}
\begin{split}
        L_{self}(\theta,\phi;\bar{\mathcal{X}}) = -log \sum \limits_{\forall \bar{x} \in \bar{\mathcal{X}}} \bar{y}(\mathcal{P}_\phi(\mathcal{F}_\theta(\bar{x})))
    \end{split}
    \label{eq:eqname2}
\end{equation}
in which $\bar{\mathcal{X}}$ is a set of randomly paired images from the training image set.

Meanwhile, we use labeled images to train the supervised pipeline with the standard cross-entropy loss. Note that the SSL and SL branch are trained in parallel from scratch.

\noindent\textbf{Self-supervised methods}\quad Inspired by \cite{c:howwell},we consider the following state-of-the-art self-supervised methods: InsDis \cite{c:insdis}, MoCo-v1 \cite{c:moco}, MoCo-v2 \cite{c:mocov2}, PIRL \cite{c:pirl}, InfoMin \cite{c:infomin} and SimSiam \cite{c:simsam}. Firstly, we download the weights of ResNet50 models pre-trained on ImageNet \cite{c:imagenet} for these SSL methods. Then, we use the standard pre-trained ResNet50 model available from the PyTorch library \cite{c:pytorch} as a baseline. The training data is fed to the backbone to attain feature vectors at first. Then we attach the classifier on top of the backbone.

Since these models are pre-trained in their own way, there may exist differences in the data augmentation methods, numbers of training epochs, training batch size, and some specific tricks that have been adopted particularly. Whereas with the same backbone and the same input image size, we can conduct a fair comparison among them. In addition, to compare with different network architectures, we make an investigation on SSL methods based on AmdimNet \cite{c:amdim} and ResNet10 \cite{c:resnet10}.
\begin{table*}
\centering
\begin{tabular}{ccccccc}
\hline
\hline
{\textbf{Methods}} & {} & {\textbf{ChestX}} & {} & {} & {\textbf{ISIC}} & {}\\
\cline{2-7}
{} & {5-way 5-shot} & {5-way 20-shot} & \multicolumn{1}{c|}{5-way 50-shot} & {5-way 5-shot} & {5-way 20-shot} & {5-way 50-shot}\\
\cline{2-7}
{\emph{MAML}} & 23.48 $\pm$ 0.96 & 27.53 $\pm$ 0.43 & \multicolumn{1}{c|}{-} & 40.13 $\pm$ 0.58 & 52.36 $\pm$ 0.57 & -\\
{\emph{ProtoNet}} & 24.05 $\pm$ 1.01 & 28.21 $\pm$ 1.15 & \multicolumn{1}{c|}{29.32 $\pm$ 1.12} & 39.57 $\pm$ 0.57 & 49.50 $\pm$ 0.55 & 51.99 $\pm$ 0.52\\
{\emph{RelationNet}} & 22.96 $\pm$ 0.88 & 26.63 $\pm$ 0.92 & \multicolumn{1}{c|}{28.45 $\pm$ 1.20} & 39.41 $\pm$ 0.58 & 41.77 $\pm$ 0.49 & 49.32 $\pm$ 0.51\\
{\emph{MetaOpt}} & 22.53 $\pm$ 0.91 & 25.53 $\pm$ 1.02 & \multicolumn{1}{c|}{29.35 $\pm$ 0.99} & 36.28 $\pm$ 0.50 & 49.42 $\pm$ 0.60 & 54.80 $\pm$ 0.54\\
\hline
{\emph{supervised}} & \underline{25.50 $\pm$ 0.41} & \underline{30.58 $\pm$ 0.44} & \multicolumn{1}{c|}{\underline{33.37 $\pm$ 0.45}} & \underline{43.13 $\pm$ 0.57} & \underline{53.61 $\pm$ 0.55} & \underline{58.70 $\pm$ 0.53}\\
{\emph{multi-task}} & 23.92 $\pm$ 0.39 & 28.51 $\pm$ 0.43 & \multicolumn{1}{c|}{30.61 $\pm$ 0.44} & 40.72 $\pm$ 0.56 & 50.18 $\pm$ 0.53 & 54.12 $\pm$ 0.56\\
{\emph{self-supervised}} & \textbf{26.80 $\pm$ 0.45} & \textbf{32.90 $\pm$ 0.47} & \multicolumn{1}{c|}{\textbf{37.05 $\pm$ 0.48}} & \textbf{43.74 $\pm$ 0.55} & \textbf{54.61 $\pm$ 0.54} & \textbf{60.86 $\pm$ 0.51}\\
\hline
\end{tabular}
\end{table*}

\begin{table*}[h]
\centering
\begin{tabular}{ccccccc}
\hline
\hline
{\textbf{Methods}} & {} & {\textbf{EuroSAT}} & {} & {} & {\textbf{CropDiseases}} & {}\\
\cline{2-7}
{} & {5-way 5-shot} & {5-way 20-shot} & \multicolumn{1}{c|}{5-way 50-shot} & {5-way 5-shot} & {5-way 20-shot} & {5-way 50-shot}\\
\cline{2-7}
{\emph{MAML}} & 71.70 $\pm$ 0.72 & 81.95 $\pm$ 0.55 & \multicolumn{1}{c|}{-} & 78.05 $\pm$ 0.68 & 89.75 $\pm$ 0.42 & -\\
{\emph{ProtoNet}} & 73.29 $\pm$ 0.71 & 82.27 $\pm$ 0.57 & \multicolumn{1}{c|}{80.48 $\pm$ 0.57} & 79.72 $\pm$ 0.67 & 88.15 $\pm$ 0.51 & 90.81 $\pm$ 0.43\\
{\emph{RelationNet}} & 61.31 $\pm$ 0.72 & 74.43 $\pm$ 0.66 & \multicolumn{1}{c|}{74.91 $\pm$ 0.58} & 68.99 $\pm$ 0.75 & 80.45 $\pm$ 0.64 & 85.08 $\pm$ 0.53\\
{\emph{MetaOpt}} & 64.44 $\pm$ 0.73 & 79.19 $\pm$ 0.62 & \multicolumn{1}{c|}{83.62 $\pm$ 0.58} & 68.41 $\pm$ 0.73 & 82.89 $\pm$ 0.54 & 91.76 $\pm$ 0.38\\
\hline
{\emph{supervised}} & \underline{78.57 $\pm$ 0.66} & \underline{85.35 $\pm$ 0.53} & \multicolumn{1}{c|}{\underline{88.94 $\pm$ 0.43}} & \underline{85.36 $\pm$ 0.60} & \underline{92.84 $\pm$ 0.38} & \underline{94.64 $\pm$ 0.32}\\
{\emph{multi-task}} & 72.41 $\pm$ 0.68 & 79.43 $\pm$ 0.58 & \multicolumn{1}{c|}{81.31 $\pm$ 0.58} & 80.41 $\pm$ 0.69 & 89.42 $\pm$ 0.47 & 92.06 $\pm$ 0.38\\
{\emph{self-supervised}} & \textbf{81.10 $\pm$ 0.62} & \textbf{88.54 $\pm$ 0.48} & \multicolumn{1}{c|}{\textbf{91.40 $\pm$ 0.39}} & \textbf{88.09 $\pm$ 0.56} & \textbf{94.95 $\pm$ 0.34} & \textbf{96.27 $\pm$ 0.29}\\
\hline
\end{tabular}
\caption{Evaluation of meta-learning methods, supervised learning method, multi-task learning method and self-supervised learning method. Average cross-domain few-shot classification accuracy (\%) with 95\% confidence intervals on the benchmark. Results style: \textbf{best} and \underline{second best}.}
\label{tab:tabname1}
\end{table*}

\begin{table}[h]
\centering
\begin{tabular}{lcccccc}
\hline
\hline
{} & {LR} & {CC} & {SVM} & {NN} & {Proto} & {LC}\\
\hline
\emph{SL} & 76.44 & 68.51 & 73.73 & 68.40 & 75.43 & 74.24 \\
\emph{MT} & \textbf{78.43} & \textbf{72.33} & \textbf{75.94} & \textbf{72.40} & \textbf{77.95} & \textbf{74.79} \\
\emph{SSL} & 62.15 & 53.12 & 60.80 & 52.57 & 60.06 & 59.51 \\
\hline
\end{tabular}
\caption{Results (\%) of Supervised method (SL), Multi-task method (MT) and Self-supervised (SSL) method trained ResNet10 models. We consider 5-way 5-shot evaluated on the test set of MiniImageNet.}
\label{tab:tabname2}
\end{table}

\begin{table*}[h]
\centering
\begin{tabular}{ccccccc}
\hline
\hline
{\textbf{Methods}} & {} & {} & {\textbf{ChestX}} & {} & {} & {}\\
\cline{2-7}
{} & {LR} & {CC} & {SVM} & {NN} & {Proto} & {LC}\\
\cline{2-7}
{\emph{SL.ResNet10}} & 25.47 $\pm$ 0.43 & 24.02 $\pm$ 0.40 & 25.44 $\pm$ 0.43 & 24.32 $\pm$ 0.40 & 25.80 $\pm$ 0.43 & 25.50 $\pm$ 0.41\\
{\emph{SSL.ResNet10}} & \underline{26.53 $\pm$ 0.45} & 24.77 $\pm$ 0.41 & 26.24 $\pm$ 0.42 & 24.42 $\pm$ 0.43 & 26.44 $\pm$ 0.46 & \textbf{26.80 $\pm$ 0.45}\\
{\emph{MT.ResNet10}} & 25.19 $\pm$ 0.42 & 23.55 $\pm$ 0.40 & 24.11 $\pm$ 0.40 & 23.86 $\pm$ 0.40 & 25.02 $\pm$ 0.42 & 23.92 $\pm$ 0.39\\
\hline
{\emph{SL.AmdimNet}} & 26.56 $\pm$ 0.45 & 24.22 $\pm$ 0.40 & 25.77 $\pm$ 0.44 & 24.46 $\pm$ 0.42 & 26.63 $\pm$ 0.45 & {26.35 $\pm$ 0.45}\\
{\emph{SSL.AmdimNet}} &  26.34 $\pm$ 0.44 & 24.89 $\pm$ 0.44 & \underline{26.76 $\pm$ 0.43} & 24.44 $\pm$ 0.38 & 26.51 $\pm$ 0.44 & \textbf{26.83 $\pm$ 0.45}\\
\hline
\hline
{\textbf{Methods}} & {} & {} & {\textbf{ISIC}} & {} & {} & {}\\
\cline{2-7}
{} & {LR} & {CC} & {SVM} & {NN} & {Proto} & {LC}\\
\cline{2-7}
{\emph{SL.ResNet10}} & 41.62 $\pm$ 0.54 & 37.71 $\pm$ 0.55 & 42.83 $\pm$ 0.58 & 38.00 $\pm$ 0.55 & 41.39 $\pm$ 0.56 & 43.13 $\pm$ 0.57\\
{\emph{SSL.ResNet10}} & 42.68 $\pm$ 0.51 & 37.32 $\pm$ 0.50 & 42.78 $\pm$ 0.55 & 37.44 $\pm$ 0.52 & 42.42 $\pm$ 0.52 & \textbf{43.74 $\pm$ 0.55}\\
{\emph{MT.ResNet10}} & \underline{43.50 $\pm$ 0.57} & 40.20 $\pm$ 0.58& 38.70 $\pm$ 0.54& 39.61 $\pm$ 0.55& 42.46 $\pm$ 0.58& 40.72 $\pm$ 0.56\\
\hline
{\emph{SL.AmdimNet}} & 43.26 $\pm$ 0.58 & 39.27 $\pm$ 0.55 & 42.01 $\pm$ 0.59 & 38.14 $\pm$ 0.55 & 41.94 $\pm$ 0.58 & 42.59 $\pm$ 0.59\\
{\emph{SSL.AmdimNet}} & \textbf{47.10 $\pm$ 0.55} & 41.05 $\pm$ 0.57 & 46.41 $\pm$ 0.57 & 41.21 $\pm$ 0.53 & \underline{46.56 $\pm$ 0.59} & 45.25 $\pm$ 0.58\\
\hline
\hline
{\textbf{Methods}} & {} & {} & {\textbf{EuroSAT}} & {} & {} & {}\\
\cline{2-7}
{} & {LR} & {CC} & {SVM} & {NN} & {Proto} & {LC}\\
\cline{2-7}
{\emph{SL.ResNet10}} & 75.08 $\pm$ 0.72 & 73.30 $\pm$ 0.78 & 76.87 $\pm$ 0.06 & 73.66 $\pm$ 0.06 & 76.86 $\pm$ 0.06 & {78.57 $\pm$ 0.66}\\
{\emph{SSL.ResNet10}} & 80.27 $\pm$ 0.63 & 75.31 $\pm$ 0.66 & 80.63 $\pm$ 0.67 & 75.48 $\pm$ 0.67 & \underline{80.97 $\pm$ 0.63} & \textbf{81.10 $\pm$ 0.62}\\
{\emph{MT.ResNet10}} & 74.90 $\pm$ 0.68 & 70.20 $\pm$ 0.75 & 66.85 $\pm$ 0.06 & 70.89 $\pm$ 0.06 & 74.99 $\pm$ 0.05 & {72.41 $\pm$ 0.68}\\
\hline
{\emph{SL.AmdimNet}} & 75.52 $\pm$ 0.68 & 71.97 $\pm$ 0.71 & 76.08 $\pm$ 0.67 & 71.94 $\pm$ 0.72 & 76.30 $\pm$ 0.66 & 75.69 $\pm$ 0.69\\
{\emph{SSL.AmdimNet}} & 82.94 $\pm$ 0.58 & 79.12 $\pm$ 0.67 &  \textbf{83.76 $\pm$ 0.59} & 79.05 $\pm$ 0.66 & \underline{83.28 $\pm$ 0.63} & 81.35 $\pm$ 0.64\\
\hline
\hline
{\textbf{Methods}} & {} & {} & {\textbf{CropDiseases}} & {} & {} & {}\\
\cline{2-7}
{} & {LR} & {CC} & {SVM} & {NN} & {Proto} & {LC}\\
\cline{2-7}
{\emph{SL.ResNet10}} & 83.96 $\pm$ 0.61 & 79.37 $\pm$ 0.66 & 84.90 $\pm$ 0.60 & 79.16 $\pm$ 0.71 & 84.13 $\pm$ 0.61 & 85.36 $\pm$ 0.60\\
{\emph{SSL.ResNet10}} & 86.28 $\pm$ 0.59 & 83.63 $\pm$ 0.62 & \textbf{88.42 $\pm$ 0.57} & 83.34 $\pm$ 0.66 & 86.90 $\pm$ 0.57 & \underline{88.09 $\pm$ 0.56}\\
{\emph{MT.ResNet10}} & 85.06 $\pm$ 0.61 & 80.52 $\pm$ 0.66 & 81.15 $\pm$ 0.64 & 80.60 $\pm$ 0.68 & 84.68 $\pm$ 0.59 & 80.41 $\pm$ 0.69\\
\hline
{\emph{SL.AmdimNet}} & 77.31 $\pm$ 0.68 & 71.09 $\pm$ 0.70 & 77.72 $\pm$ 0.64 & 69.75 $\pm$ 0.71 & 76.01 $\pm$ 0.67 & 77.08 $\pm$ 0.68\\
{\emph{SSL.AmdimNet}} & \underline{91.24 $\pm$ 0.48} & 88.98 $\pm$ 0.53 & \textbf{92.12 $\pm$ 0.46} & 88.58 $\pm$ 0.53 & 90.72 $\pm$ 0.52 & 87.38 $\pm$ 0.61\\
\hline
\end{tabular}
\caption{Results of supervised method, multi-task method and self-supervised method trained models evaluated by various classifiers. ResNet10 and AmdimNet are used in this experiment. Results style: \textbf{best} and \underline{second best} for each architecture.}
\label{tab:tabname3}
\end{table*}

\begin{figure*}[t]
    \centering
    \centerline{\includegraphics[width=17cm]{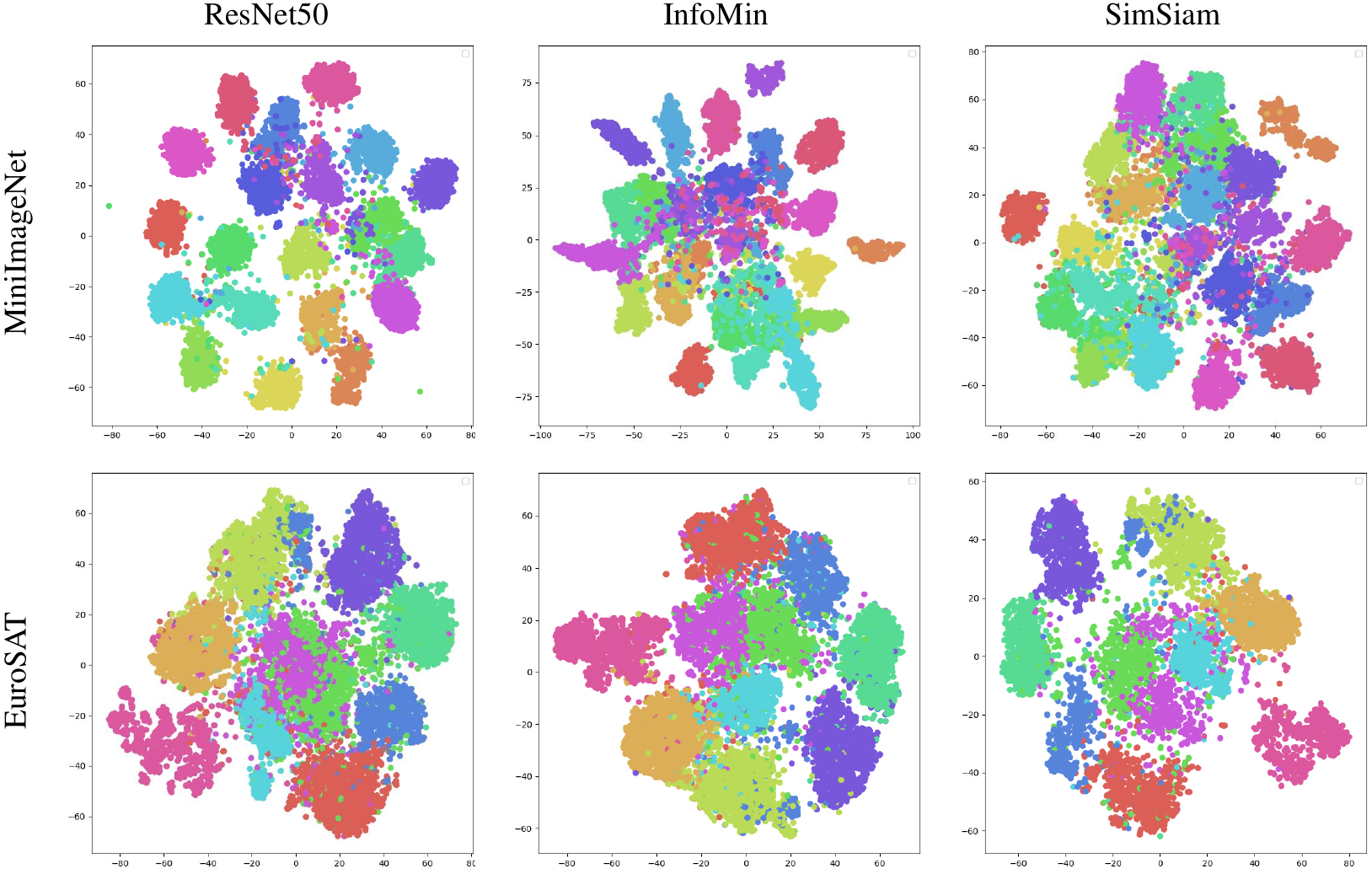}}
    \caption{t-SNE plots of trained embeddings on 20 novel classes from MiniImageNet and 10 classes from EuroSAT, one color represents one class. Training methods considered are the supervised learned ResNet50, InfoMin and SimSiam. All are trained from ImageNet and tested respectively on MiniImageNet and EuroSAT.}
    \label{fig:tsne}
\end{figure*}

\section{Experiments}
In this section, we begin by introducing the benchmarks used in our work. Specifically, we execute quantitative experiments and carry out an adequate study to clarify our motivation, thereby answering the critical questions we asked in detail. All experiments are conducted on a single NVIDIA GeForce GTX 2080Ti GPU with 11GB memory.

\subsection{Benchmarks}
To train embedding networks, we use MiniImageNet dataset \cite{c:match}, which is a subset of ImageNet containing 60,000 images from 100 classes, each with 600 images. Following the data split strategy in \cite{c:optim}, we sample 64 classes as the training set, 16 classes as the validation set, and 20 classes as the test set. The test set of MiniImageNet is used to evaluate the in-domain performance of the learned representations.

Suggested by \cite{c:broader}, 4 datasets are used as benchmarks to evaluate the generalization of the embedding networks. 1) CropDiseases \cite{c:cd}, natural images specific to agriculture industry, 2) EuroSAT \cite{c:eurosat}, satellite images from a bird's eye view, 3) ISIC \cite{c:isic}, medical skin image under unnatural scenes, 4) ChestX \cite{c:chestx}, X-ray chest images as grayscale images. The four datasets exhibit decreasing similarity to the MiniImageNet.

\subsection{To what extent self-supervised features can transfer directly to uncharted domains?}
\textbf{Experimental Setup}\quad We train the supervised learning method, multi-task method, and self-supervised learning method on the base classes of MiniImageNet. Specifically, they are trained with the same augmentation methods including random crop, resize, colorjitter, random horizontal flip, and normalization. As mentioned in \cite{c:sslk}, architecture choices may significantly affect the performance of self-supervised methods, we reproduce representation networks with the same architecture for each approach in particular. Each method is performed on ResNet10 with input image size $224\times224$.

We adopt the recommended hyper-parameters in \cite{c:rfs} for the supervised method, \cite{c:bf3s} for the multi-task method and \cite{c:prototrans} for the self-supervised method. Specifically, we choose the multi-task learning method with the pretext task of rotation prediction since it performs significantly better than location prediction. For meta-learning methods, the model is meta-trained on MiniImageNet for 400 epochs with Adam optimizer. The initial learning rate is set to 0.001.

During evaluation, we comply with the same 600 randomly sampled episodes through a random Numpy seed 10 (for consistency) to train a Linear Classifier. For each task, the query set always has 15 images per class and the Linear Classifier is trained for 100 epochs using SGD with momentum. The learning rate is 0.01 and the momentum rate is 0.9. We adopt the cross-entropy loss to train it. In these experiments, we consider 5-way 5-shot, 5-way 20-shot, and 5-way 50-shot settings. If not specified, backbones will be frozen during the evaluation of all experiments.

\noindent\textbf{Results}\quad As reported in Table \ref{tab:tabname1}, the SSL method dominates in all target datasets under all settings and exceeds the supervised method by around $2\%$ on average, illustrating that SSL with a shallow network can achieve good performance as well. The performance of the multi-task method is somehow not satisfying. Though it achieves the best scores in MiniImageNet under a 5-way 5-shot setting as presented in Table \ref{tab:tabname2}, it suffers from the over-fitting problem to the source domain just like the meta-learning method. In contrast, the SSL method performs badly in MiniImageNet but shows brilliant transferability to the target domains. Intuitively, an embedding that performs well in-domain has no direct relation with its performance cross-domain. Note that existing methods usually use novel classes from the source domain as the validation set, such that the model with the best transferability to novel tasks in-domain will be selected, which is not optimal for cross-domain verification.

\subsection{A good embedding is all you need?}
\noindent\textbf{Experimental Setup}\quad Another line of work exploits a broad variety of classifiers to test the performance of feature encoders on CDFSL. Besides Linear Classifier, we implement 5 other well-known classifiers to further validate the performance of each method. We record the average accuracy and 95\% confidence interval of each method tested on the target dataset. The 5-way 5-shot transfer is used in all experiments for consistency if no additional instructions. Moreover, we consider the most recent self-supervised method used in CDFSL called AmdimNet, which is a deeper and more complex architecture compared to ResNet10. We evaluate the performance of AmdimNet respectively trained by self-supervised and supervised learning following the same hyper-parameter settings as \cite{c:amdim}.

\noindent\textbf{Results}\quad From Table \ref{tab:tabname3}, we observe that if the embedding performs well in one classifier, it can also perform well in other classifiers. This is in line with the general view that good embedding is the basis for good performance when transferring to a new task. Nevertheless, the performance with each method varies a lot among different classifiers, e.g. the performance of SSL by ResNet10 has a gap of 5.79\% among classifiers in EuroSAT, with 5.27\% for supervised method and 8.14\% for the multi-task method. This phenomenon also exists in other datasets. Surly the choice of classifiers can have a great impact on the results, which shows that some recent work about metric learning on FSL is of great significance. Moreover, We found that the best score in each dataset is often obtained by LC. As the only method that needs to be trained in the feed-forward network, the time-consuming disadvantage can be exchanged for good performance.

For the AmdimNet backbone, we compare the performance of the supervised and self-supervised methods trained on MiniImageNet and tested on the target domains in Table \ref{tab:tabname3}. Compared with ResNet10, the SSL in AmdimNet exceeds the supervised method to a greater extent, indicating that a deeper architecture can stimulate the potential of SSL.
\subsection{Which one is the most practically effective among the plethora of self-supervised methods?}
\noindent\textbf{Experimental Setup}\quad Given numerous SSL methods with state-of-the-art performance, we apply 6 outstanding SSL methods among them. A ResNet50 model trained from the supervised way is set as the baseline. Our evaluation uses various classifiers on the features extracted from the ResNet50 backbones with the input image size $224\times224$.

\noindent\textbf{Results}\quad By thoroughly evaluating our large suite of recent SSL methods on transfer to the CDFSL task, Table \ref{tab:tabname4} shows that the most advanced SSL method performs better than the baseline in most cases. With the similarity of the target domain to MiniImageNet (source domain) decreasing, SSL methods achieve better generalization to the target domain in general. The baseline gets the best score of 92.65\% in CropDiseases. However, it consistently lags behind the most advanced SSL method in other datasets, which conveys that SSL methods can preserve higher transferable knowledge under more acute domain shifts. Importantly, \textbf{the ranking of methods is not coherent across different domains}, denoting prevalently applicable SSL methods are still vacant. Combining the results from Table \ref{tab:tabname3} and Table \ref{tab:tabname4}, we find that SSL show greater enhancement under smaller training data.

\begin{table*}[h]
\fontsize{6.4pt}{8pt}
\selectfont
\centering
\begin{tabular}{ccccccccccc}
\hline
\hline
{\textbf{Methods}} & {} & {}  & {\textbf{ChestX}} & {} & {} & {} & {} & {\textbf{ISIC}} & {} & {} \\
\cline{2-11}
{} & {LR} & {CC} & {SVM} & {NN} & \multicolumn{1}{c|}{Proto} & {LR} & {CC} & {SVM} & {NN} & {Proto}\\
\cline{2-11}
{\emph{InsDis}} & 25.30 $\pm$ 0.42 & 24.40 $\pm$ 0.41 & 24.29 $\pm$ 0.40 & 23.97 $\pm$ 0.42 & \multicolumn{1}{c|}{25.31 $\pm$ 0.44} & 44.98 $\pm$ 0.57 & 39.38 $\pm$ 0.57 & 39.65 $\pm$ 0.53 & 39.57 $\pm$ 0.54 & 43.77 $\pm$ 0.57\\
{\emph{PIRL}} & 25.94 $\pm$ 0.43 & 24.17 $\pm$ 0.40 & 24.96 $\pm$ 0.43 & 24.09 $\pm$ 0.40 & \multicolumn{1}{c|}{25.78 $\pm$ 0.43} & \underline{46.05 $\pm$ 0.56} & \textbf{41.08 $\pm$ 0.53} & \underline{41.70 $\pm$ 0.56} & \textbf{40.70 $\pm$ 0.54} & \underline{45.28 $\pm$ 0.55}\\
{\emph{MoCo-v1}} & \underline{26.17 $\pm$ 0.44} & \underline{24.77 $\pm$ 0.42} & 24.34 $\pm$ 0.42 & \underline{24.62 $\pm$ 0.41} & \multicolumn{1}{c|}{\underline{26.00 $\pm$ 0.43}} & \textbf{46.07 $\pm$ 0.57} & 40.70 $\pm$ 0.55 & 38.50 $\pm$ 0.51 & 40.20 $\pm$ 0.54 & \textbf{45.34 $\pm$ 0.58}\\
{\emph{MoCo-v2}} & 24.88 $\pm$ 0.42 & 23.57 $\pm$ 0.41 & 25.15 $\pm$ 0.45 & 23.98 $\pm$ 0.41 & \multicolumn{1}{c|}{24.98 $\pm$ 0.42} & 45.26 $\pm$ 0.55 & 39.60 $\pm$ 0.53 & 41.30 $\pm$ 0.54 & \underline{40.26 $\pm$ 0.54} & 43.98 $\pm$ 0.57\\
{\emph{InfoMin}} & 24.37 $\pm$ 0.42 & 23.32 $\pm$ 0.40 & 24.71 $\pm$ 0.41 & 23.46 $\pm$ 0.42 & \multicolumn{1}{c|}{24.50 $\pm$ 0.43} & 41.84 $\pm$ 0.56 & 37.84 $\pm$ 0.54 & 37.67 $\pm$ 0.55 & 37.71 $\pm$ 0.54 & 40.73 $\pm$ 0.57\\
{\emph{SimSiam}} & \textbf{26.75 $\pm$ 0.43} & \textbf{24.87 $\pm$ 0.42} & \textbf{26.00 $\pm$ 0.41} & \textbf{25.02 $\pm$ 0.43} & \multicolumn{1}{c|}{\textbf{26.67 $\pm$ 0.44}} & 45.37 $\pm$ 0.56 & \underline{40.75 $\pm$ 0.54} & 41.66 $\pm$ 0.55 & 40.24 $\pm$ 0.56 & 44.94 $\pm$ 0.58\\
{\emph{Supervised}} & 25.90 $\pm$ 0.46 & 23.71 $\pm$ 0.41 & \underline{25.46 $\pm$ 0.44} & 23.78 $\pm$ 0.43 & \multicolumn{1}{c|}{25.28 $\pm$ 0.45} & 41.96 $\pm$ 0.55 & 38.38 $\pm$ 0.55 & \textbf{43.84 $\pm$ 0.57} & 37.76 $\pm$ 0.54 & 40.85 $\pm$ 0.57\\
\hline
{\emph{Speed\_avg}} & {3.35} & \textbf{8.35} & {7.61} & {7.60} & \multicolumn{1}{c|}{\underline{8.15}} & {2.13} & {7.71} & \underline{6.63} & {6.37} & \textbf{6.87}\\
\hline
\end{tabular}
\end{table*}

\begin{table*}[h]
\fontsize{6.4pt}{8pt}
\selectfont
\centering
\begin{tabular}{ccccccccccc}
\hline
\hline
{\textbf{Methods}} & {} & {} & {\textbf{EuroSAT}} & {} & {} & {} & {} & {\textbf{CropDiseases}} & {} & {}\\
\cline{2-11}
{} & {LR} & {CC} & {SVM} & {NN} & \multicolumn{1}{c|}{Proto} & {LR} & {CC} & {SVM} & {NN} & {Proto}\\
\cline{2-11}
{\emph{InsDis}} & 82.04 $\pm$ 0.58 & 79.20 $\pm$ 0.66 & 75.47 $\pm$ 0.70 & 78.91 $\pm$ 0.68 & \multicolumn{1}{c|}{81.90 $\pm$ 0.63} & 88.43 $\pm$ 0.57 & 88.33 $\pm$ 0.57 & 85.97 $\pm$ 0.66 & 88.61 $\pm$ 0.56 & 88.74 $\pm$ 0.57\\
{\emph{PIRL}} & 82.12 $\pm$ 0.62 & 79.96 $\pm$ 0.65 & 77.10 $\pm$ 0.66 & 79.90 $\pm$ 0.67 & \multicolumn{1}{c|}{83.06 $\pm$ 0.62} & 88.61 $\pm$ 0.60 & 88.09 $\pm$ 0.59 & 87.24 $\pm$ 0.58 & 88.36 $\pm$ 0.57 & 88.07 $\pm$ 0.58\\
{\emph{MoCo-v1}} & 82.26 $\pm$ 0.64 & 79.56 $\pm$ 0.64 & 74.07 $\pm$ 0.67 & 78.98 $\pm$ 0.70 & \multicolumn{1}{c|}{82.49 $\pm$ 0.65} & \underline{90.02 $\pm$ 0.51} & 88.35 $\pm$ 0.58 & 81.34 $\pm$ 0.66 & 88.71 $\pm$ 0.55 & 88.62 $\pm$ 0.55\\
{\emph{MoCo-v2}} & 85.83 $\pm$ 0.52 & 81.22 $\pm$ 0.60 & 82.59 $\pm$ 0.56 & 80.70 $\pm$ 0.61 & \multicolumn{1}{c|}{85.34 $\pm$ 0.55} & 89.63 $\pm$ 0.56 & 88.27 $\pm$ 0.64 & \underline{90.25 $\pm$ 0.51} & 87.67 $\pm$ 0.54 & 89.81 $\pm$ 0.60\\
{\emph{InfoMin}} & \textbf{86.35 $\pm$ 0.45} & \textbf{82.44 $\pm$ 0.54} & 79.99 $\pm$ 0.54 & \textbf{82.54 $\pm$ 0.54} & \multicolumn{1}{c|}{\underline{86.31 $\pm$ 0.47}} & 89.16 $\pm$ 0.60 & 88.67 $\pm$ 0.58 & 89.62 $\pm$ 0.57 & 88.71 $\pm$ 0.58 & 89.68 $\pm$ 0.56\\
{\emph{SimSiam}} & \underline{86.24 $\pm$ 0.50} & \underline{82.12 $\pm$ 0.56} & \underline{84.14 $\pm$ 0.52} & \underline{82.50 $\pm$ 0.55} & \multicolumn{1}{c|}{\textbf{86.50 $\pm$ 0.47}} & \textbf{91.04 $\pm$ 0.53} & \textbf{89.99 $\pm$ 0.54} & 88.94 $\pm$ 0.57 & \textbf{90.32 $\pm$ 0.53} & \textbf{91.15 $\pm$ 0.53}\\
{\emph{Supervised}} & 85.00 $\pm$ 0.53 & 81.54 $\pm$ 0.60 & \textbf{84.59 $\pm$ 0.51} & 81.64 $\pm$ 0.58 & \multicolumn{1}{c|}{85.41 $\pm$ 0.55} & 89.97 $\pm$ 0.53 & \underline{88.77 $\pm$ 0.57} & \textbf{92.65 $\pm$ 0.47} & \underline{88.78 $\pm$ 0.57} & \underline{90.46 $\pm$ 0.51}\\
\hline
{\emph{Speed\_avg}} & {2.09} & \textbf{6.31} & {5.12} & {4.99} & \multicolumn{1}{c|}{\underline{6.18}} & {1.78} & {6.21} & {6.10} & \underline{6.95} & \textbf{8.03}\\
\hline
\end{tabular}
\caption{Comparison between SOTA self-supervised methods with the standard supervised model under a broad array of classifiers and the average speed (number of iteration per second it/s) for each classifier. Results style: \textbf{best} and \underline{second best}.}
\label{tab:tabname4}
\end{table*}

\subsection{Discussion}
\noindent\textbf{Why do SSL methods perform well in CDFSL?}\quad The supervised method learns compact representations by narrowing the intra-category discrepancy while learning discriminative features, as depicted in Figure \ref{fig:tsne}: each novel class in MiniImageNet can be clustered tightly in the latent feature space. On the contrary, features encoded by SimSiam in each class are perceptually looser and more difficult to separate correctly. However, we observe that under sharp domain shifts, the distribution of encoded features moves in the opposite way. Features cast by SimSiam are more distinguishable than the supervised way in EuroSAT. The key difference remains in the design of the loss function. The intra-class invariance enforced by the traditional supervised loss weakens transferability and causes task misalignment. We believe that widening inter-class distance while retaining intra-class discriminates contributes to improved uncertainty calibration to new tasks.

\begin{figure}[t]
    \centerline{\includegraphics[width=8cm]{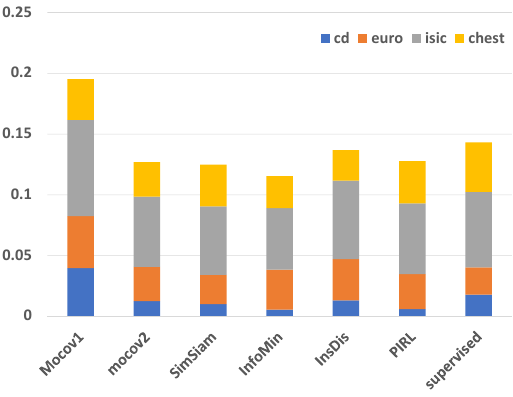}}
    \caption{The Coefficient of Variation (C.V) is calculated by $ {\sigma \over \mu }$, where $\sigma$ defines the standard deviation among classifiers, $\mu$ stands for the average results of classifiers. Image is plotted from results listed in Table \ref{tab:tabname4}.}
    \label{fig:cv}
\end{figure}

Meanwhile, we calculate respectively the Coefficient of Variation (C.V) of the accuracy for each method under different classifiers, the results are shown in Figure \ref{fig:cv}. A low C.V means a smoothly good performance of one method under various classifiers. InfoMin presents steady performance compared to other methods, except in EuroSAT the SimSiam method behaves the best. Most self-supervised methods produce more robust representations than the supervised method, whereas MoCo-v1 is relatively unstable.

\noindent\textbf{Which classifier is the best and how we ideally measure their performance?}

\begin{figure}[t]
    \includegraphics[width=8.4cm]{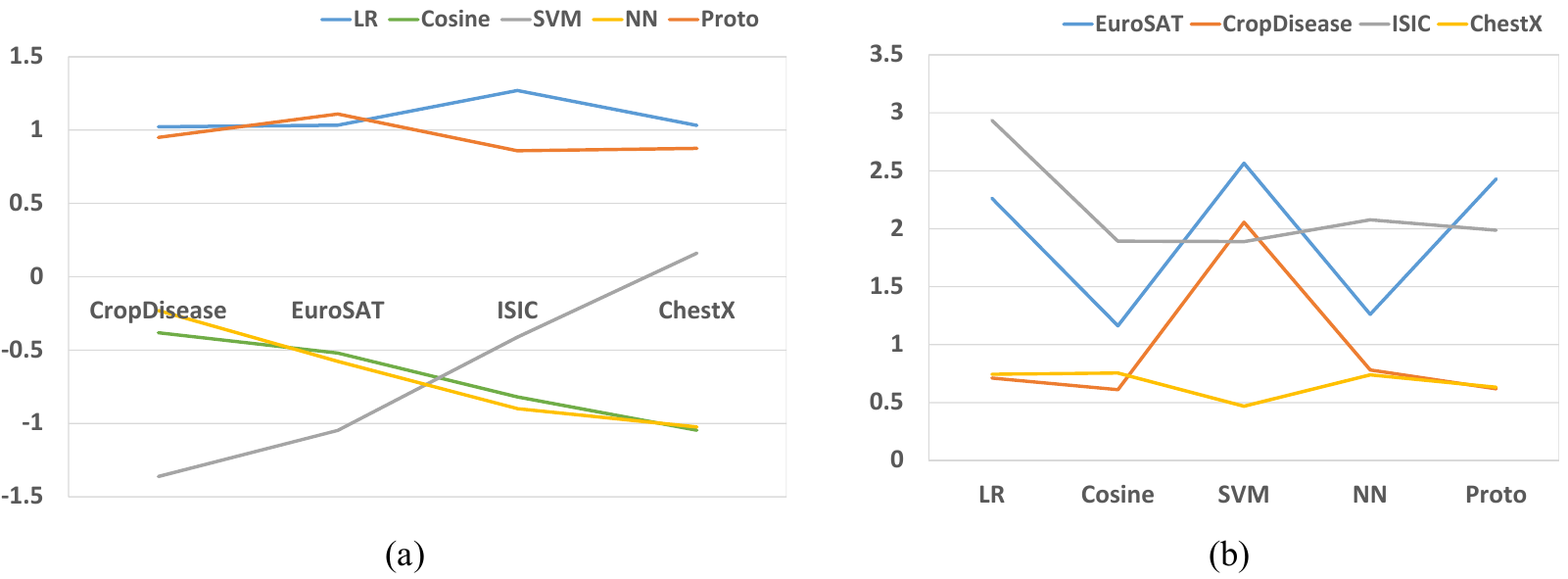}
    \caption{(a) the distance between the performance (\%) of each classifier with their averaged performance; (b) The Z-Score of the averaged performance of each classifier among all classifiers. The performance of each classifier is firstly averaged by all methods in each dataset listed in Table \ref{tab:tabname4}.}
    \label{fig:zscore}
\end{figure}

\noindent We analyze each classifier in three aspects:

1) \textit{Capability}, Figure \ref{fig:zscore} (a) shows that both Linear Regression and Prototypical Classifier are above average, indicating their good classification ability in general. The performance of Cosine Classifier and Nearest Neighbor declines rapidly as the difficulty of the task increases (the target dataset is dissimilar to the source dataset more).

2) \textit{Stability}, good classifiers should have stable performance under various conditions, so as to fairly reflect the quality of each embedding. We utilize Z-Score $z$ as a metric of stability. A lower $z$ stands for better stability. Cosine Classifier and Prototypical Classifier show promising stability across 4 datasets as illustrated in Figure \ref{fig:zscore} (b). The trend of each depicted line conforms to the data distribution of each dataset, in which CropDiseases and EuroSAT are relatively similar, while ISIC and ChestX are comparatively similar.

3) \textit{Velocity}, the average speed of each classifier under different methods in each benchmark is recorded in Table \ref{tab:tabname4}. Prototypical Classifier and Cosine Classifier show excellent average speed in each dataset.

In summary, Prototypical Classifier externalizes outstanding performance, competitive stability, and prominent speed compared to other classifiers. Note that Linear Classifier is eliminated here due to its inefficiency: it is sensitive to the learning schedule and requires many epochs to converge. Therefore, we recommend Prototypical Classifier as the standard evaluation recipe for CDFSL.

\section{Conclusion}
In this paper, we delve into a large empirical benchmarking study on the efficacy of pre-trained embeddings for CDFSL. The remarkable generalization of SSL methods that has been speculated for a long time is now clearly confirmed on this downstream task. Notably, existing methods use novel classes from the source domain to validate and select the best model during training, which is not optimal and resulting in over-fitting to the source domain. This suggests that an appropriate validation way under cross-domain settings remains to be realized. Besides, we neglect the transductive learning paradigm where testing data can be leveraged to boost performance. This is to say, the research on the domain-specific self-supervised representations trained from each target domain is left to future work.

\noindent\textbf{Acknowledgements}: This work was supported in part by the National Natural Science Foundation of China (No. 62106236).

\bibliography{mybib, articles}

\end{document}